\documentclass[runningheads]{llncs}
\usepackage[T1]{fontenc}
\usepackage{algorithmic}
\usepackage[ruled,vlined]{algorithm2e}  
\usepackage{graphicx}
\usepackage{amsfonts}
\newcommand{\colorpm}[1]{_{\footnotesize{\pm#1}}}
\usepackage[colorlinks]{hyperref}

\usepackage{lipsum} 

\newcommand\blfootnote[1]{%
  \begingroup
  \renewcommand\thefootnote{}\footnote{#1}%
  \addtocounter{footnote}{-1}%
  \endgroup
}

\usepackage[misc]{ifsym}

\begin{document}
\title{Robust Remote Sensing Scene Classification with Multi-View Voting and Entropy Ranking}
%
%
\author{Jinyang Wang\inst{1,2} 
\and Tao Wang\inst{1,2^{(\href{mailto:twang@mju.edu.cn}{\textrm{\scriptsize\Letter}})}}
\and Min Gan\inst{2^{(\href{mailto:aganmin@aliyun.com}{\textrm{\scriptsize\Letter}})}}
\and George Hadjichristofi\inst{3}
}
\authorrunning{J. Wang et al.}
%
\institute{
Fujian Provincial Key Laboratory of Information Processing and Intelligent Control,
College of Computer and Control Engineering and International Digital Economy College, Minjiang University, Fuzhou 350108, China\\
\and
College of Computer and Data Science, Fuzhou University, Fuzhou 350108, China\\
\and
Department of Computer Science and Engineering, European University Cyprus, Nicosia 1516, Cyprus\\
\blfootnote{Paper accepted by the 4th International Conference on Machine Learning for Cyber Security (ML4CS 2022), Guangzhou, China. The final authenticated version is available online at \url{https://doi.org/10.1007/978-3-031-20096-0_7}. }
}
\maketitle              
\begin{abstract}
Deep convolutional neural networks have been widely used in scene classification of remotely sensed images. In this work, we propose a robust learning method for the task that is secure against partially incorrect categorization of images. Specifically, we remove and correct errors in the labels progressively by iterative multi-view voting and entropy ranking. At each time step, we first divide the training data into disjoint parts for separate training and voting. The unanimity in the voting reveals the correctness of the labels, so that we can train a strong model with only the images with unanimous votes. In addition, we adopt entropy as an effective measure for prediction uncertainty, in order to partially recover labeling errors by ranking and selection. We empirically demonstrate the superiority of the proposed method on the WHU-RS19 dataset and the AID dataset.

\end{abstract}
\section{Introduction}

Remotely sensed imagery is an important means to acquire information about the Earth with wide applications in geography, land surveying, ecology, and oceanography, among others. Particularly, remote sensing (RS) scene classification is a fundamental task in RS image analytics that classifies each image into a predefined set of semantic categories. As large-scale RS image datasets become more easily accessible, it has recently been feasible to train effective Deep Convolutional Neural Networks (DCNNs) for the task.

One of the key limitations of training DCNNs for RS scene classification, however, is that a large amount of manual labeling is required, in order to train models with good generalization abilities. Commercial imaging companies are collecting hundreds of terabytes of new satellite images every day; it is virtually impossible to manually label even a fraction of the ever-expanding image data. One possible remedy to this dilemma would be the use of labels obtained through crowd-sourcing platforms. However, it would be difficult to eliminate labeling errors even with label voting or aggregation. In addition, labels obtained through other channels such as automated labeling methods or auxiliary information, such as user-generated labels could also be error-prone, as manual verification of the labels is time-consuming. In this paper, we address the problem of training DCNNs on datasets with noisy labels. Here, labels being ``noisy'' means that some of the images in the datasets are incorrectly categorized. Conventionally, training DCNNs with noisy labels would inevitably lead to a rapid degradation of model performance. From a model learning perspective, it would be ideal if we could train models that are robust to label noise. Importantly, this would suggest that the learning algorithms are more secure against deliberated attacks such as label manipulation, which is a vital consideration in the security of machine learning algorithms.

In this work, we propose a robust remote sensing scene classification method with iterative multi-view voting and entropy ranking. Our method is able to deal with noisy labels by progressively removing and rectifying labeling errors. Specifically, at each time step, we first divide the training data into disjoint parts for separate training and voting. The unanimity in the voting reveals the correctness of the labels, so that we can train a strong model with only the images with unanimous votes. In addition, we adopt entropy as an effective measure for prediction uncertainty, in order to partially recover labeling errors by ranking and selection. By performing the above steps repeatedly, the quality of the labeling and the trained models is gradually improved. The main contributions of our method are two-fold: 1) We propose a simple yet effective method for robust scene classification of remotely sensed images. The two main steps, multi-view voting and entropy ranking, are complementary and enhance each other. Specifically, multi-view voting provides a way to identify errors in labels, and entropy ranking could then partially rectify these errors based on prediction uncertainty. 2) Our method provides a general framework for robust learning, and works easily with different deep models. Also, experiments on two public datasets, WHU-RS19 and AID, demonstrate the efficacy of the proposed method.

The rest of the paper is organized as follows. Sec.~\ref{sec:related} briefly reviews recent work on deep learning-based image classification with noisy labels, as well as those for remote sensing scene classification. Sec.~\ref{sec:approach} describes the proposed method in detail, followed by experimental evaluation results in Sec.~\ref{sec:exp} and closing remarks in Sec.~\ref{sec:conclusion}.

\section{Related Work}
\label{sec:related}

Many deep learning-based methods have been proposed to solve the problem of label noise in generic image classification. For example, Wang $et\ al.$~\cite{wang2019symmetric} proved that on the ``simple'' classes, the cross-entropy learning criterion will overfit noisy labels. To address this issue, the authors proposed anti-noise reverse cross-entropy, a method of symmetric cross-entropy learning. Xia $et\ al.$~\cite{xia2020robust} approached the problem by making a distinction between critical and non-critical parameters. The effect of noisy labels is scaled down by applying different rules for these two types of parameters. Another closely related work from Kaneko $et\ al.$~\cite{kaneko2019label} proposed label-noise robust Generative Adversarial Networks (GANs), which trained a noise transition model to extract knowledge from clean labels even in the presence of noise. In addition, Huang $et\ al.$~\cite{huang2019o2u} proposed O2U-net to detect noisy labels through parameter adjustments. Specifically, the network can be transferred from the over-fitting state to the under-fitting state cyclically, during which the normalized average loss of a sample is recorded for the identification of label noise. Vahdat~\cite{vahdat2017toward} proposed a conditional random field that represents the relationship between noisy and clean labels in a semi-supervised setting. Furthermore, Cordeiro $et\ al.$~\cite{cordeiro2021longremix} divided the training data into clean and noisy sets in an unsupervised manner, followed by semi-supervised learning to minimize the empirical vicinal risk (EVR). They demonstrated the importance of the accuracy of the unsupervised classifier and the the size of the training set to minimize the EVR. Consequently, a new algorithm, termed LongReMix, was proposed. Unlike existing work, in this paper we propose to refine noisy labels in an iterative fashion by multi-view voting and entropy ranking.

In the realm of RS image classification, there have also been some recent attempts to address the problem of label noise. For instance, Kang $et\ al.$~\cite{kang2020robust} used the negative Box–Cox transformation to downplay the effect of noisy labels, and proposed a robust normalized softmax loss. Li $et\ al.$~\cite{li2020error} used multi-view CNNs to correct errors in labels, and proposed an error-tolerant deep learning method, in addition to an adaptive multi-feature collaborative representation classifier to improve classification quality. Tu $et\ al.$~\cite{tu2020robust} designed a covariance matrix representation-based noisy label model. Li $et\ al.$~\cite{li2021complementary} combined ordinary learning and complementary learning as the latter can reduce the probability of learning misinformation. Being closest to our work, Li $et\ al.$~\cite{li2019learning} proposed a robust scene classification method based on multi-view voting. Our work differs from theirs as we use entropy as an important additional predictor for label correctness. In particular, we propose a way to rectify the labels of images with non-unanimous votes, which is not possible with their method. Also, the entropy ranking method we proposed provides performance improvements in addition to multi-view voting at different label noise ratios on two public datasets.



\begin{figure}[htbp]  
\centering  
\includegraphics[width=\textwidth]{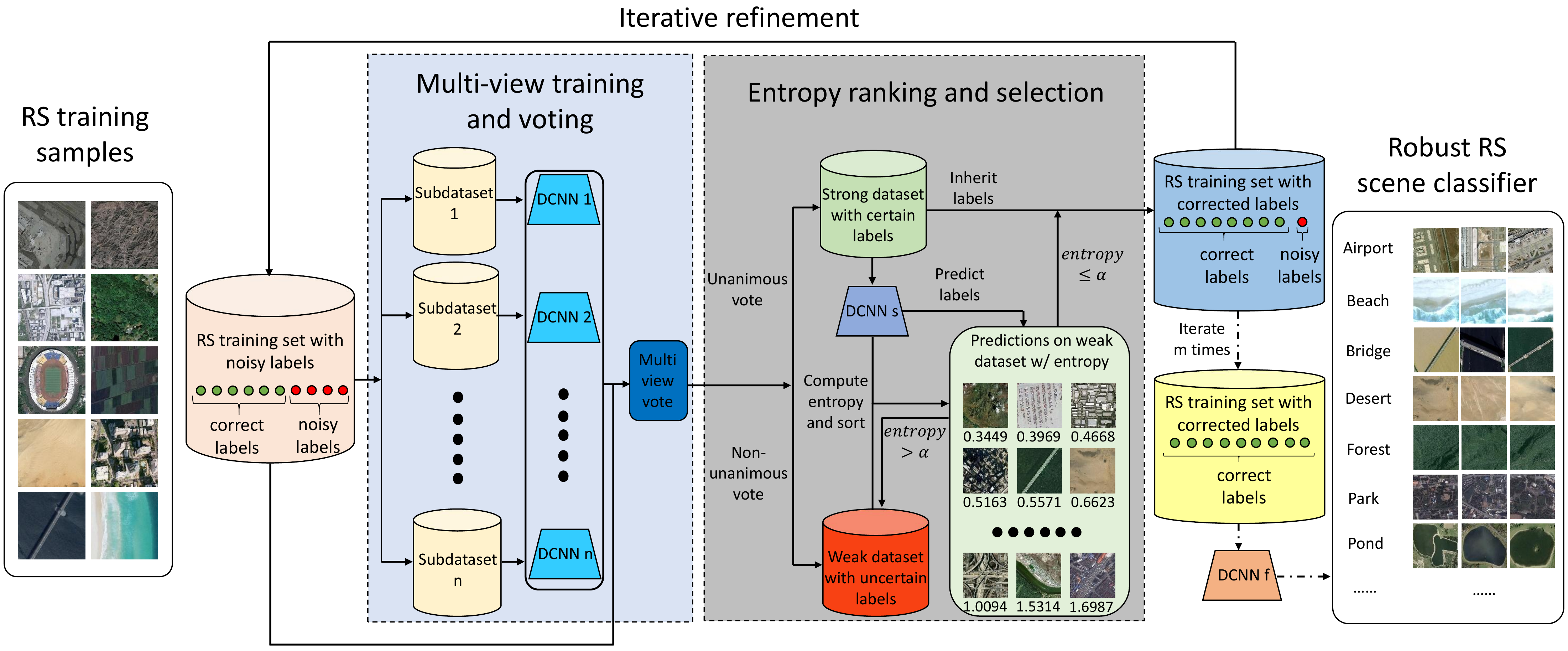}  \caption{Robust remote sensing scene classification via iterative multi-view voting and entropy ranking. In each iteration, the dataset with noisy labels is divided into disjoint subsets for multi-view training and voting (shown in the blue rectangle). Based on voting unanimity, the dataset is split into a strong dataset with certain labels and a weak dataset with uncertain labels. A DCNN is trained with the strong dataset and based on its prediction entropy, confident samples with lower entropy are further added into the strong dataset (shown in the gray rectangle). The above steps are carried out several times to produce a final training set. See Sec.~\ref{sec:approach} for details.} 
\label{fig:overall}
\end{figure}

\section{Our Approach}
\label{sec:approach}

In this paper, we propose an approach to RS scene classification that is robust to label noise. Given an initial training set with noisy labels, we perform iterative refinement via multi-view voting and entropy ranking to produce a cleaner dataset with fewer incorrect labels. Our method is based on two key observations: 1) images with correct labels tend to receive unanimous votes when we split the original dataset into multiple subsets to perform multi-view voting, and 2) the entropy of predictions can be used as an additional predictor for label correctness. Thus, we refine our dataset in an iterative fashion with both multi-view voting and entropy ranking. The overall process is illustrated in Fig.~\ref{fig:overall}.

More specifically, we take two steps for dataset refinement in each iteration: 1) multi-view training and voting (shown in the blue rectangle in Fig.~\ref{fig:overall}), and 2) entropy ranking and selection (shown in the gray rectangle in Fig.~\ref{fig:overall}). All the cylinders in Fig.~\ref{fig:overall} represent training datasets, the green dots in the cylinders represent correctly labeled images, and the red dots represent incorrectly labeled images.
In the multi-view training and voting step, we split the training set into $n$ disjoint subsets to train $n$ DCNNs. Based on the predictions from these DCNNs, the original dataset with noisy labels can be divided into a strong dataset and a weak dataset. The samples in the strong dataset are those with consistent votes across all $n$ DCNN models, and the weak dataset contains samples with inconsistent voting results.
In the entropy ranking and selection step, a portion of noisy labels is converted into correct labels. This is done by training a DCNN on the strong dataset and evaluating its prediction entropy on the weak dataset. Here, entropy is used as an effective measure of label correctness. Samples with lower prediction entropy (with their labels converted according to the most likely class in prediction) are further added into the strong dataset, to produce a cleaner output dataset for the current iteration. After $M$ iterations, a dataset with the lowest proportion of incorrect labels will be obtained, and a DCNN is trained with this dataset to obtain a robust classification model. We summarize the main procedure of our algorithm in Algorithm 1.

The rest of this section is organized as follows. In Subsections~\ref{sec:mvv} and ~\ref{sec:er}, we describe details in the multi-view training and voting step and the entropy ranking step, respectively. Subsection~\ref{sec:ir} provides details on the iterative refinement.



\RestyleAlgo{ruled}
   \begin{algorithm}[t]
     \caption{The proposed algorithm with multi-view voting and entropy ranking.}

   \KwIn{Dataset with noisy labels $\mathcal{D}$. Semantic classes $\mathcal{C}$.}
    Iterative refinement with $M$ iterations:
    
   \For{ $m = 1 : M$ }
   {

     Divide the dataset $\mathcal{D}_m$ ($\mathcal{D}_1 = \mathcal{D}$. $\mathcal{D}_m$ = $\mathcal{D}^s_{m-1}$ if $m>1$) into $n$ disjoint parts:\\
     $\mathcal{S}_1\cap\mathcal{S}_2\cap\cdot\cdot\cdot\cap\mathcal{S}_n$=$\emptyset$,\ \ \ 
     $\mathcal{D}$=$\mathcal{S}_1\cup\mathcal{S}_2\cup\cdot\cdot\cdot\cup\mathcal{S}_n$
     
     \For{ $\mathcal{S}_j = \mathcal{S}_1 : \mathcal{S}_n$ }
     {
        Train a Deep Convolutional Neural Network~(DCNN) with $\mathcal{S}_j$\\
        \Return $\mathit{DCNN}_j$
     }
     
     \For{  $\mathit{DCNN}_t$ =  $\mathit{DCNN}_1$ :  $\mathit{DCNN}_n$ }
     {
        \For{ $(\mathbf{x}_i,y_i)$ in $\mathcal{D}_m$ }
        {
            Predict the label $z_i^t = \arg \max_{c \in \mathcal{C}}p_{\mathit{DCNN}_t}(\mathbf{x}_i=c)$ for each image
        }
     }
     \For{ $(\mathbf{x}_i,y_i)$ in $\mathcal{D}_m$ }
        {
            If $z_i^1$=$z_i^2$=$\cdot\cdot\cdot$=$z_i^n$: 
            label the image accordingly and place the image into strong dataset $ (\mathbf{x}_i,  y_i) \rightarrow \mathcal{D}^s$\\
            Else : place the image into weak dataset $\mathbf{x}_i \rightarrow \mathcal{D}^w$
        }
     Train strong ${\textit{DCNN}}_s$ with $\mathcal{D}^s$ \\
     \For{ $\mathbf{x}_i$ in $\mathcal{D}^w$ }
        {
           Calculate the entropy of $E^w(\mathbf{x}_i)$\\
           If $E^w(\mathbf{x}_i) \leq \alpha$ : predict label $\hat{y}_i$ with ${\textit{DCNN}}_s$ and place the image into strong dataset $(\mathbf{x}_i, \hat{y_i}) \rightarrow \mathcal{D}^s$ \\
        }
     Output dataset for the current iteration $\mathcal{D}^s_m = \mathcal{D}^s$
   }
    Train a final model $\textit{DCNN}_f$ with $\mathcal{D}^s_M$ \\
    \Return ${DCNN}_f$
    
   \label{alg:pseudolabel}
   \end{algorithm}

\subsection{Multi-View Training and Voting}
\label{sec:mvv}

For datasets containing noisy labels, it has been shown that the negative impact of noisy labels can be effectively mitigated by splitting the dataset into several parts and training them separately. Here we follow the multi-view training and voting method outlined in~\cite{li2019learning} while removing the training of a post-hoc Support Vector Machine (SVM)~\cite{vapnik1999nature} for simplicity considerations.

More formally, denote the input RS image as $\mathbf{x} \in \mathbb{R}^{H \times W \times 3}$ and $y \in \mathcal{C}$ its corresponding semantic class label, where $\mathcal{C}=\{1 \dots C\}$ is the label space. At the beginning of training, we are given an input dataset $\mathcal{D} = \{(\mathbf{x}_i, y_i) \}_{i=1}^N$ whose labels may be partially incorrect. In addition, denote $\mathcal{D}_m$ as the input training set at the $m$-th iteration of our algorithm. The total number of iterations is $M$ (see Subsection~\ref{sec:ir}). Naturally we have $\mathcal{D}_1=\mathcal{D}$.
Apart from $\mathcal{D}_m$, the iteration number $m$ is omitted below for notation simplicity. In each iteration, we split $\mathcal{D}_m$ into $n$ disjoint parts, i.e., $\mathcal{S}_1\cap\mathcal{S}_2\cap\cdot\cdot\cdot\cap\mathcal{S}_n$=$\emptyset$,\ \ \ 
     $\mathcal{D}_m$=$\mathcal{S}_1\cup\mathcal{S}_2\cup\cdot\cdot\cdot\cup\mathcal{S}_n$.
       Next, $n$ DCNNs are trained on $\mathcal{S}_1, \mathcal{S}_2, \cdots \mathcal{S}_n$, respectively, so as to obtain $\mathit{DCNN}_1$, $\mathit{DCNN}_2$, $\cdot\cdot\cdot$, $\mathit{DCNN}_n$. We use these DCNN models to make predictions on all RS images in $\mathcal{D}_m$, and if the prediction results of all models for a given image $\mathbf{x}$ are consistent, the image is placed in the strong
       dataset $\mathcal{D}^s = \{(\mathbf{x}_i,y_i)\}_{i=1}^K$. Otherwise, we remove the label corresponding to this image and place it into the weak
       dataset $\mathcal{D}^w = \{\mathbf{x}_i\}_{i=K+1}^N$.
       We note that, once an image is placed into the weak dataset $\mathcal{D}^w$, it shall remain
       in $\mathcal{D}^w$ across iterations such that the number of images in $\mathcal{D}^s$ and
       $\mathcal{D}^w$ should always add up to $N$. However, as we will show in Sec~\ref{sec:er},
       the strong dataset $\mathcal{D}^s$ and consequently the quality of the model trained on $\mathcal{D}^s$ will improve over time. Therefore, images in $\mathcal{D}^w$ may be moved back to $\mathcal{D}^s$ at a later time.
       In the above, $\mathbf{x}_i$ represents the $i$-th image in the strong dataset, $y_i$ represents the predicted label of the $i$-th image, $N$ represents the number of images in the entire training set.
    
\begin{figure}[htbp]  
\centering  
\includegraphics[width=\textwidth]{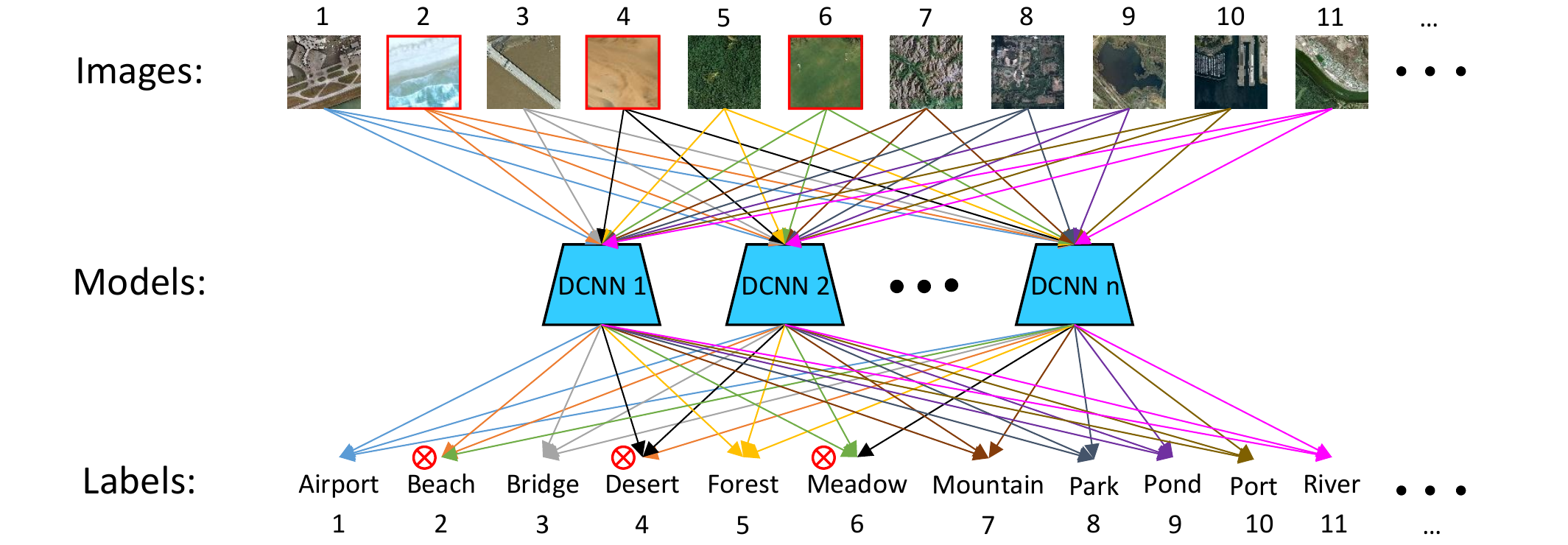}  
\caption{Example of the multi-view voting process. All images are jointly predicted by all DCNN models. Depending on whether the votes are unanimous or non-unanimous, we sort the images into a strong dataset (images with unanimous votes) and a weak dataset (images with non-unanimous votes, marked with red borders and a cross next to the corresponding labels). Lines with different colors represent the training and prediction paths for different training images. See text for details.}
\label{fig:vote}
\end{figure}

    Fig.~\ref{fig:vote} illustrates the voting process. In the top row and the bottom row are images and their corresponding labels. The trapezoidal blocks in the middle represent the trained DCNN models after splitting the dataset. The arrows between the images and the DCNNs indicates that each image will get predictions from all DCNNs, and the arrows between the DCNNs and the labels indicate the predictions by voting. If the votes are unanimous, then the image will be added into the strong dataset. Otherwise, the corresponding image will be added into the weak dataset (images shown with red border in the top row, and also with a cross by their corresponding labels in the bottom row).
    
\subsection{Entropy Ranking}
\label{sec:er}

After we obtain $\mathcal{D}^s$ with the steps outlined in the previous section, we can train a strong DCNN model $\mathit{DCNN}_s$ with $\mathcal{D}^s$. We note that this DCNN will likely be trained with cleaner labels, so that its predictions will less likely be affected by label noise. In this regard, we further evaluate the predictions of $\mathit{DCNN}_s$ on the weak dataset $\mathcal{D}^w$, in the hope that we can additionally recover some images from $\mathcal{D}^w$. Here, we use entropy as a proxy for prediction certainty, i.e., we only choose images receiving a low entropy in the prediction by $\mathit{DCNN}_s$.

Specifically, given an image from the weak dataset $\mathbf{x}_i \in \mathcal{D}^w$, denote $p(\mathbf{x}_i=c)$ as the the output probability for the $c$-th class by $\mathit{DCNN}_s$, the entropy $E^w(\mathbf{x}_i)$ can be written as:
    
    
\begin{equation}
    E^w(\mathbf{x}_i)=-\sum\limits_{c \in \mathcal{C}}p(\mathbf{x}_i=c)\log p(\mathbf{x}_i=c)
\end{equation}

We then sort the entropy for all images in $\mathcal{D}^w$ in ascending order, and take a hyperparameters $\alpha$ as the entropy threshold. Specifically, we move the image from the weak dataset into the strong dataset if the prediction entropy of an image is not greater than $\alpha$. If $\alpha$ is lower, we only move images with high prediction certainty from the weak dataset to the strong dataset, potentially missing some images with correct predictions as a result. If $\alpha$ is higher, we move more images into the strong dataset, resulting in a higher chance of incorrect labels in the strong dataset. In practice, the value of $\alpha$ is chosen empirically (see discussion in Subsection~\ref{sec:id}). The process above can be written as:

\begin{equation}
    \mathcal{D}^w = \mathcal{D}^w \backslash \mathbf{x}_i, \: (\mathbf{x}_i, \hat{y}_i) \rightarrow \mathcal{D}^s, \quad \textrm{if} \: E^w(\mathbf{x}_i) \leq \alpha
\end{equation}

\noindent where $\mathcal{D}^w = \mathcal{D}^w \backslash \mathbf{x}_i$ means that we remove $\mathbf{x}_i$ from $\mathcal{D}^w$ and $(\mathbf{x}_i, \hat{y}_i) \rightarrow \mathcal{D}^s$ indicates that we add $(\mathbf{x}_i, \hat{y}_i)$ into $\mathcal{D}^s$. Here, $\hat{y}_i = \arg \max_{c \in \mathcal{C}} p(\mathbf{x}_i=c)$ is the maximum a posteriori (MAP) estimate provided by $\mathit{DCNN}_s$, as the truth label $y_i$ may be noisy and hence not used.
    

\subsection{Iterative Refinement}
\label{sec:ir}

In practice, the multi-view voting results may be partially incorrect and that noisy labels could sometimes be included in the strong dataset $\mathcal{D}^s$. This problem can be alleviated by performing the steps in Sec.~\ref{sec:mvv} and Sec.~\ref{sec:er} in an iterative fashion and gradually improving the quality of the strong dataset. Specifically, denote $\mathcal{D}^s_{m-1}$ as the strong dataset we obtained at the end of the previous iteration. We let this strong dataset be the input dataset in the next iteration, i.e., $\mathcal{D}_{m} = \mathcal{D}^s_{m-1}$.
In our experiments, we found that the iterative refinement of $\mathcal{D}^s$ is particularly important when the proportion of noisy labels is large. We set the total number of iterations to $M=3$ empirically as it provides the largest performance gain in most cases.

\section{Experiments and Discussions}
\label{sec:exp}

In this section, we use two widely used public RS image datasets to demonstrate the efficacy of the proposed noise-tolerant robust RS scene classification method. In the following, we first describe the details of the datasets and our implementation, and then present results and discussions.

\subsection{Datasets}

    \noindent \textbf{WHU-RS19 dataset~\cite{sheng2012high}.} The WHU-RS19 dataset has a total of $1,005$ images, which belong to $19$ different classes. The size of each image is 600 $\times$ 600 pixels. Following~\cite{li2019learning}, we use $50\%$ of the images for training and the remaining $50\%$ for testing.
    
    \noindent \textbf{Aerial Image Datasets (AID) dataset~\cite{xia2017aid}.} The AID dataset has a total of $10,000$ images, which belong to $30$ different classes. The number of images per class is between $200$ and $500$, and the size of each image is 600 $\times$ 600 pixels. Again, we use $50\%$ of the images for training and the remaining $50\%$ for testing following~\cite{li2019learning}.
    
\subsection{Implementation Details}
\label{sec:id}

    In our experiments, we introduce random errors from $10\%$ to $50\%$ (at $10\%$ interval) of the original correct labels, in order to create varying levels of label noise. We use a 50-layer ResNet~\cite{he2016deep} pretrained on ImageNet~\cite{deng2009imagenet} for all DCNNs. In the preprocessing step, all images are resized to 256 $\times$ 256 pixels.
    We use Adam~\cite{kingma2014adam} as the optimizer with learning rate set to $0.001$.
    Each DCNN is trained for 50 epochs with the standard cross-entropy loss.
    We implement our method with PyTorch, and all the experiments were run on a single NVIDIA GeForce RTX 2080 Ti GPU.
    
\begin{figure}[htbp]
    \centering
    \begin{minipage}[t]{0.45\textwidth}
    \centering
    \includegraphics[width=5.5cm]{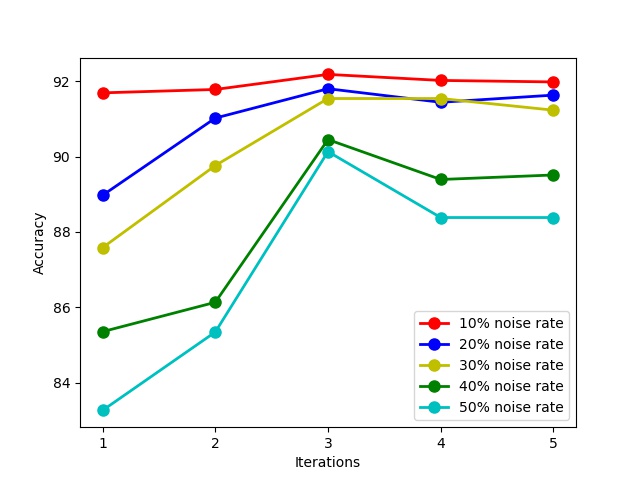}
    \caption{Classification accuracy \textit{vs.} number of iterations $M$ on the WHU-RS19 dataset under varying label noise ratios.}
    \label{fig:numiter}
    \end{minipage}
    \begin{minipage}[t]{0.1\textwidth}
    \end{minipage}
    \begin{minipage}[t]{0.45\textwidth}
    \centering
    \includegraphics[width=5.7cm]{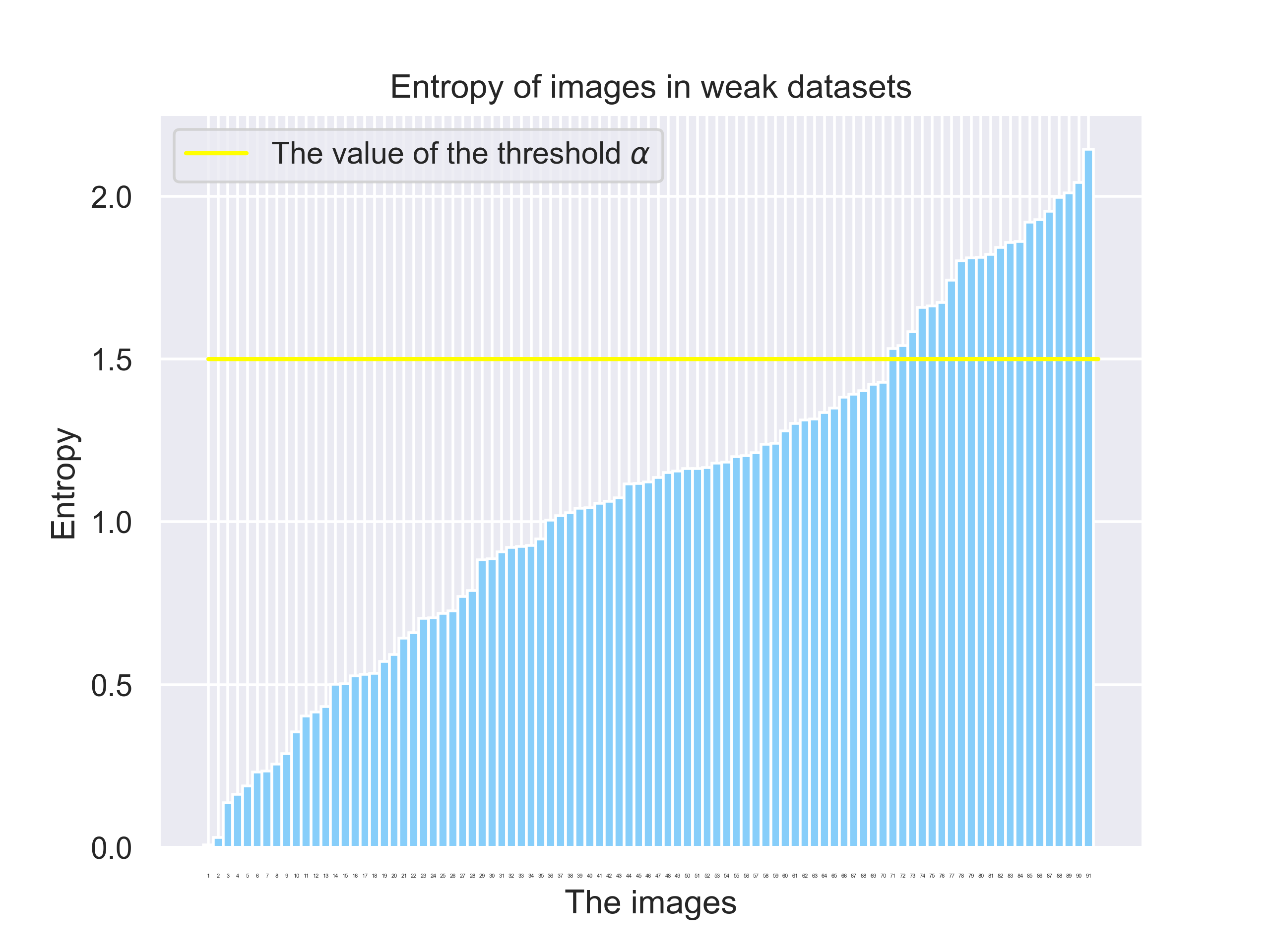}
    \caption{An example entropy distribution in the weak dataset on the WHU-RS19 dataset ($40\%$ noise ratio, $2$nd iteration).}
    \label{fig:entropy}
    \end{minipage}
    \end{figure}

    In order to ensure fairness of the experiments, in the initial training set, the proportion of label noise in each class is consistent with the proportion of label noise in the entire training set, and the noisy labels are randomly distributed. 
    The hyperparameter $M$, which is the number of iterations, is chosen empirically. Specifically, we perform experiments with different noise ratios, i.e., $10\%$, $20\%$, $30\%$, $40\%$, and $50\%$. For each noise ratio, we perform $10$ different noise distribution experiments.  Following~\cite{li2019learning}, two DCNN views(i.e., $n=2$) are used to validate our method, and we will try more cases in subsequent works. Finally, the average of the results is taken to choose the number of iterations $M$. We present the average accuracy results we obtained in Fig.~\ref{fig:numiter}. It is clear that $M=3$ works best and we use this value throughout our experiments. In particular, increasing $M$ beyond $3$ may lead to over-fitting to the label noise so it may even lead to a degradation of performance.
    Another important hyperparameter $\alpha$, which is the entropy threshold for recovering images from the weak dataset, is chosen by grid search. We present an example in Fig.~\ref{fig:entropy} that shows a typical value of $1.5$ for $\alpha$ on the WHU-RS19 dataset.

\subsection{Results}
We present the quantitative results of our experiments in Table~\ref{tab:exp_whu} and Table~\ref{tab:exp_aid}. Table~\ref{tab:exp_whu} shows the experimental evaluation results on the WHU-RS19 dataset. Table~\ref{tab:exp_aid} shows the experimental evaluation results on the AID dataset.
It is clear that our method, Remote Sensing Multi-View Voting and Entropy Ranking (RS-MVVER), is superior to the competing method by a clear margin. We note that while the accuracy improvements may seem small in some cases, our method provides consistent improvements at all noise ratios. In addition, even a small improvement is important for remote sensing scene classification which may involve images of critical infrastructures, and we generally want the results to be as accurate as possible.  Specifically, we compare our method against two baseline methods: ResNet-50~\cite{he2016deep} and RS-ETDL~\cite{li2019learning}. We note that the latter is a strong baseline method that also uses multi-view voting for learning from noisy labels in remote sensing scene classification. The results suggest that both multi-view voting and entropy ranking can positively impact model performance.

\begin{table}[htbp]
\centering
\caption{ Overall accuracy(\%) on WHU-RS19 using different ratios of noisy labels. The average accuracy and the $95\%$ confidence interval across $10$ experiments are reported.}
\begin{tabular}{|c|c|c|c|}
         \hline
         Noise ratio &  ResNet50 in \cite{he2016deep} & RS-ETDL in \cite{li2019learning} & RS-MVVER~(Ours)\\
         \hline
         10\% &  91.35$\colorpm{1.42}$ & 91.79$\colorpm{2.22}$ & {\bfseries 92.18}$\colorpm{0.72}$\\
         20\% &  88.65$\colorpm{1.36}$ & 89.74$\colorpm{1.88}$ & {\bfseries 91.80}$\colorpm{1.38}$\\
         30\% &  87.12$\colorpm{2.63}$ & 88.91$\colorpm{1.96}$ & {\bfseries 91.53}$\colorpm{0.31}$\\
         40\% &  83.27$\colorpm{1.58}$ & 88.71$\colorpm{1.45}$ & {\bfseries 90.45}$\colorpm{1.24}$\\
         50\% &  82.12$\colorpm{3.14}$ & 89.10$\colorpm{3.57}$ & {\bfseries 90.13}$\colorpm{3.09}$\\
         \hline
\end{tabular}
\label{tab:exp_whu}
\end{table}

\begin{table}[htbp]
\centering
\caption{ Overall accuracy(\%) on AID using different ratios of noisy labels. The average accuracy and the $95\%$ confidence interval across $5$ experiments are reported.}
\begin{tabular}{|c|c|c|c|}
         \hline
         Noise ratio &  ResNet50 in \cite{he2016deep} & RS-ETDL in \cite{li2019learning} & RS-MVVER~(Ours)\\
         \hline
         10\% &  75.45$\colorpm{0.37}$ & 75.78$\colorpm{0.32}$ & {\bfseries 76.34}$\colorpm{0.38}$\\
         20\% &  71.72$\colorpm{0.31}$ & 73.93$\colorpm{0.54}$ & {\bfseries 74.60}$\colorpm{1.00}$\\
         30\% &  68.89$\colorpm{0.68}$ & 71.40$\colorpm{0.85}$ & {\bfseries 72.97}$\colorpm{0.92}$\\
         40\% &  66.39$\colorpm{0.39}$ & 70.34$\colorpm{0.23}$ & {\bfseries 70.46}$\colorpm{0.23}$\\
         50\% &  62.37$\colorpm{0.72}$ & 65.12$\colorpm{0.56}$ & {\bfseries 66.07}$\colorpm{1.33}$\\
         \hline
\end{tabular}
\label{tab:exp_aid}
\end{table}

\begin{figure}[htbp]  
    \centering  
    \includegraphics[width=\textwidth]{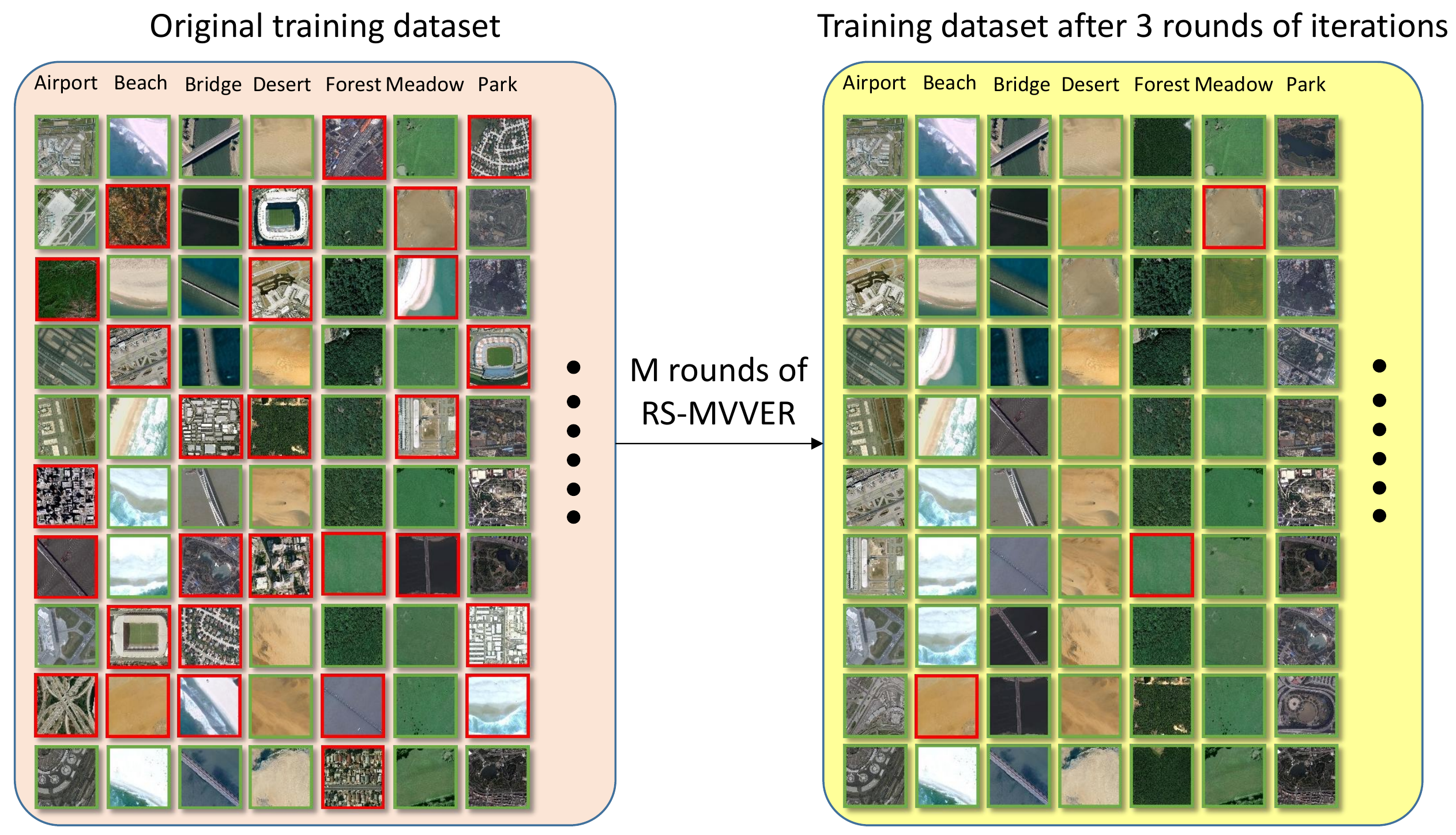}  
    \caption{Dataset before and after $M$ rounds of iterative refinement~($40\%$ noise ratio on the WHU-RS19 dataset). Images with red borders represent the wrongly labeled ones in the original training data set, and images with green borders represent the correctly labeled ones. After $3$ rounds of iterations, we can clearly see that the noisy labels are greatly reduced.} 
    \label{fig:iterrefine}
    \end{figure}

Finally, we present a qualitative example of the iterative refinement to the dataset. Fig.~\ref{fig:iterrefine} shows an example of the iterative refinement process of the WHU-RS19 dataset. The original dataset has $40\%$ noisy labels, and the number of noisy labels are either eliminated or greatly reduced for most classes after $3$ rounds of iterative refinement. We note that the iterative refinement is performed during training, so that it does not impact evaluation performance. In our experiments, a single training iteration takes $15$ and $60$ minutes on the WHU-RS19 and the AID dataset, respectively. This is acceptable as the training is performed offline.

\section{Conclusion}
\label{sec:conclusion}
In this paper, we propose a robust remote sensing scene classification method with multi-view voting and entropy ranking that can deal with training datasets containing noisy labels. Our method begins with splitting the original training dataset into disjoint parts and training a set of multi-view classifiers. Next, we use multi-view voting followed by entropy ranking to sort images into a strong dataset and a weak dataset. By iteratively perfoming the steps above, we can obtain a cleaner dataset with much fewer noisy labels. Experiments on two public datasets, WHU-RS19 and AID, show that our method compares favorably with competing methods. We hope that our method could provide some new insights and a better starting point for future work on this topic. In particular, we would like to explore incorporating knowledge from images in the weak dataset or additional unlabeled images into our learning objectives.

\subsubsection{Acknowledgment.} 
This work is supported by NSFC (61703195), Fujian NSF (2022J011112) and Fuzhou Technology Planning Program (2021-ZD-284).

%
%
%
\bibliographystyle{splncs04}
\bibliography{main}

\end{document}